%% file: _main_RSS_ROAR.tex
\documentclass[conference]{IEEEtran}
\usepackage{times}

\usepackage[numbers]{natbib}
\usepackage{multicol}
\usepackage[bookmarks=true]{hyperref}
\usepackage{graphicx}
\usepackage{wrapfig}
\usepackage{caption}
\usepackage{amsmath}
\usepackage{xcolor}
\usepackage{subfigure}
\usepackage{algorithm} 
\usepackage{algpseudocode} 
\usepackage[normalem]{ulem}

\begin{document}

\title{Learning to Predict Mobile Robot Stability in Off-Road Environments}

\author{Nathaniel Rose$^*$, Arif Ahmed$^*$, Emanuel Gutierrez-Cornejo and Parikshit Maini\\
Department of Computer Science and Engineering\\
University of Nevada Reno\\
Email: \{nathanielrose, arifa, egutierrezcornejo, pmaini\}@unr.edu}
\def\thefootnote{*}



%

\maketitle
\footnotetext{These authors contributed equally to this work.}

\input{abstract}

\input{1_Introduction}
\input{2_Related_Works}
\input{3_Solution_Approach}
\input{4_Data_Collection_and_Processing}
\input{5_Results}
\input{6_Conclusion_And_FutureWork}
\input{acknowledgement}
\bibliographystyle{unsrt}
\bibliography{references}

\end{document}

%% file: abstract.tex
\begin{abstract}
Navigating in off-road environments for wheeled mobile robots is challenging due to dynamic and rugged terrain. Traditional physics-based stability metrics, such as Static Stability Margin (SSM) or Zero Moment Point (ZMP) require knowledge of contact forces, terrain geometry, and the robot’s precise center-of-mass that are difficult to measure accurately in real-world field conditions. In this work, we propose a learning-based approach to estimate robot platform stability directly from proprioceptive data using a lightweight neural network, IMUnet. Our method enables data-driven inference of robot stability without requiring an explicit terrain model or force sensing.

We also develop a novel vision-based ArUco tracking method to compute a scalar score to quantify robot platform stability called C3 score. The score captures image-space perturbations over time as a proxy for physical instability and is used as a training signal for the neural network based model. As a pilot study, we evaluate our approach on data collected across multiple terrain types and speeds and demonstrate generalization to previously unseen conditions. These initial results highlight the potential of using IMU and robot velocity as inputs to estimate platform stability. The proposed method finds application in gating robot tasks such as precision actuation and sensing, especially for mobile manipulation tasks in agricultural and space applications. Our learning method also provides a supervision mechanism for perception based traversability estimation and planning.

\end{abstract}

%% file: 1_Introduction.tex
\section{Introduction}

Wheeled unmanned ground vehicles (UGVs) operating in unstructured, off-road environments must navigate uneven, deformable and discontinuous terrain while maintaining operational safety and performance. Stability is a critical requirement during sustained deployments in agricultural fields, forests, and space exploration, as terrain-induced disturbances can increase rollover risk, degrade perception accuracy, and interfere with task execution. These effects are especially pronounced in scenarios involving mobile manipulation or eye-in-hand sensing, where platform instability while not catastrophic can still compromise data quality and actuation precision. Physics-based stability metrics such as Static Stability Margin (SSM), Zero Moment Point (ZMP), and Stability Moment (SM) \cite{SSM, ZMP, ESM} offer rigorous definitions grounded in robot dynamics. However, their application to real-world UGVs is limited by the need to accurately determine terrain contact points, center-of-mass and measure real-time force experienced by the robot. The requirements are difficult to satisfy, especially for robot chassis with wheel suspensions and require specialized force sensors to accurately measure instantaneous force on the wheels. Learning-based stability estimation using terrain models, camera and/or lidar data that estimate the effect of the forces on robot stability indirectly, suffer from the lack of a universally accepted easy to measure metric for ground truth labeling. To train a model to recognize instability, we must first define stability in a measurable and repeatable way. Alternatively, IMU-based signal filtering methods (e.g., angular rate thresholds or standard deviation filters) are lightweight but difficult to interpret semantically and often require task-specific tuning to remain effective across varying terrain or vehicle configurations. 

\begin{figure}
    \centering
    \includegraphics[width=0.6\linewidth]{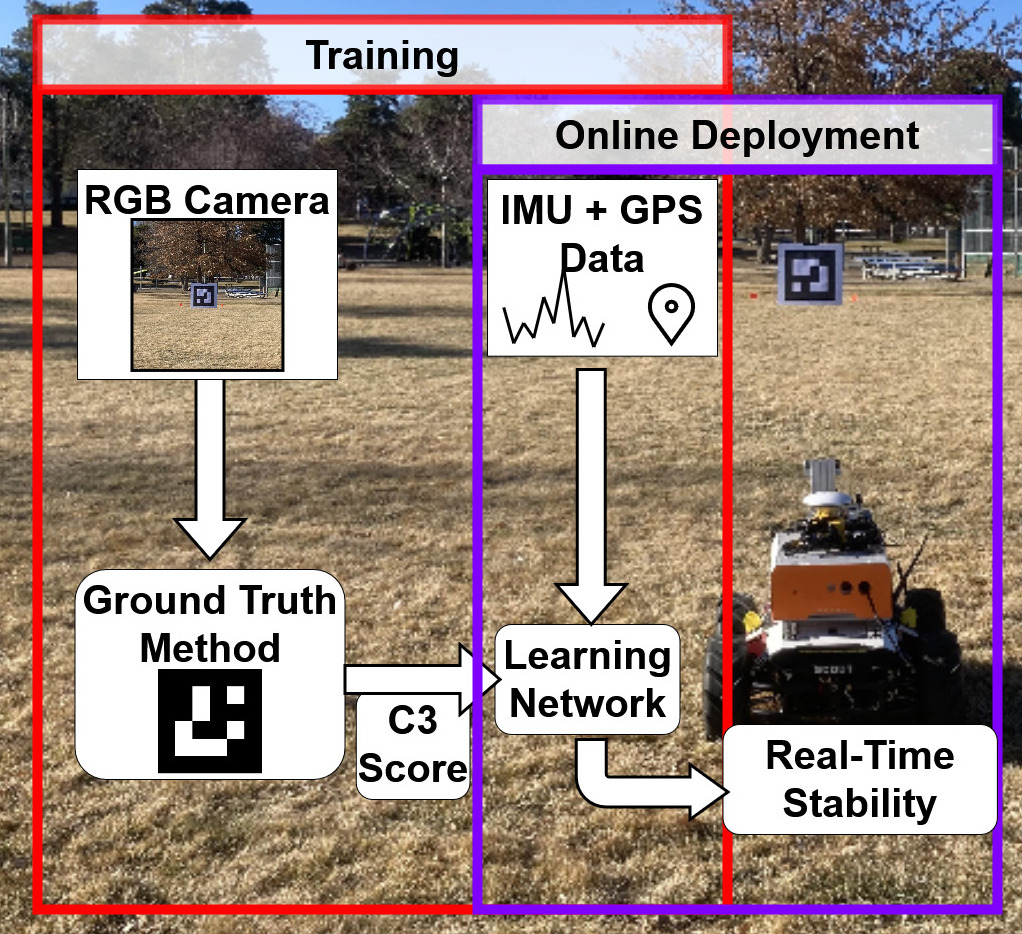}
    \caption{Proposed architecture for stability estimation}
    \label{fig:arch}
\end{figure}

In this paper, we report results from a pilot study to address these challenge by proposing a vision-supervised learning framework for stability estimation using IMU and robot velocity data. We introduce a novel stability measure, C3 score, which quantifies platform disturbance based on pose deviation captured by a static monocular camera observing ArUco markers. The C3 score combines rotational and translational displacement to characterize the magnitude of body-level motion, and is used as a training label for a neural network that maps IMU and velocity inputs to a real-time stability estimate without any reliance on vision at deployment time. This approach inverts the common near-to-far paradigm \cite{mayuku2021self}, where proprioceptive signals are typically used to supervise long-range exteroceptive terrain models. Instead, we leverage vision during training to supervise proprioceptive learning, offering a scalable, low-cost way to generate labels without force sensors or simulation (Figure \ref{fig:arch} shows the proposed pipeline for stability estimation). Once trained, the model serves as a semantic stability estimator that is both deployable in real-world systems and applicable across multiple downstream use cases. These include gating sensing or actuation in mobile manipulation (e.g., only capturing data during stable motion), retrospective terrain labeling, and self-supervised training of traversability prediction models. Key contributions of this paper are as follows. First, we define a novel vision-supervised stability metric (C3) based on pose deviation using ArUco markers. Second, we develop a learning-based pipeline that predicts this metric using only IMU and velocity data, enabling real-time proprioceptive stability estimation. Third, we evaluate the approach in real-world field experiments across multiple terrain types and discuss its practical implications for autonomous mobility and control. 

%% file: 2_Related_Works.tex
\section{Related Works}

Understanding and quantifying vehicle stability is essential for ensuring safe and reliable navigation in unstructured or uncertain terrain. In this work, we focus exclusively on methods that employ proprioceptive sensing for real-time stability estimation. These methods fall broadly into two categories: physics-driven and learning-based approaches. Physics-driven methods leverage knowledge of vehicle dynamics, forces, and moments to derive stability estimates based on analytical models. In contrast, learning-based methods rely on data-driven techniques to infer stability or traversability directly from machine learning models by evaluating on-board sensor measurements. The following sections review the key developments in both categories, highlighting their capabilities, limitations, and relevance to our proposed method.

\subsection{Physics-driven Stability Metrics}
Physics-driven stability metrics are often divided into 2 categories: static and dynamic stability. Many physics-driven metrics are proposed from the perspective of analyzing the stability of legged robots, as these robotic platforms are naturally more prone to tipping over and thus benefit the most from stability analysis, but all of the following metrics can also be effectively applied to wheeled robots, as demonstrated by Ryu et al. \cite{ryu2024evaluation, ryu2024simulation}.

\subsubsection{Static Stability Metrics} Static stability attempts to quantify a robot platform's ability to maintain its balance while stationary or moving at slow speeds. The static stability margin (SSM) metric \cite{SSM} calculates the point where a gravitational force vector, starting at the robot's center of mass (CoM) point, intersects the ground plane and the center point of the vehicle's support polygon, projected on the ground plane. The distance between these 2 points is directly proportional to the likelihood that the platform will tip over. Another metric, known as the Zero Moment Point (ZMP) metric \cite{ZMP}, uses a point projected on the ground plane where the sum of all forces and moments acting on a robot are in equilibrium, meaning the robot’s body will not rotate around the ZMP. If the ZMP falls outside of a stable region, the robot will topple. The Energy Stability Margin (ESM) metric \cite{ESM} quantifies the minimal amount of energy required to tip over a mobile robot by computing the difference in height between the maximal height of the CoM and the height of the current pivot point of the robot. 

\subsubsection{Dynamic Stability Metrics} Dynamic stability attempts to quantify the balance of a robot in motion over time when subjected to perturbations or disturbances in motion. To achieve this, dynamic stability metrics often consider how a robot responds to sudden changes in speed, direction, and external forces. The Force-Angle Stability Margin (FASM) metric \cite{FASM} determines a quantifiable stability measurement of a robotic platform based on the net force vector and moments applied to CoM. The FASM metric states that the stability of the platform will be inversely proportional to the angle between the tip-over axis normal and the net force vector components parallel to that tip-over axis. The Moment-Height Stability (MHS) metric \cite{MHS} measures dynamic stability while considering the moving CoM of a mobile manipulator. The MHS metric can be expressed as the ratio of the tipping moment over the maximal resistive moment, which is inversely proportional to the vehicle's stability. The stability moment (SM) metric \cite{SM} aggregates individual wheel contact forces and moments to quantify the risk of rollover during the operation of a wheeled vehicle at high-speed. 


Physics-driven stability metrics have proven to be useful and effective when applicable; however, these methods can be impractical to implement for online UGV deployment. Many physics-driven stability metrics require localization of the robot wheels' points of contact on the terrain relative to the UGV's CoM, and/or estimation of real-time forces applied on the wheels of the UGV. Norouzi et al. \cite{norouzi2013statistical} further document that many off-road terrain scenarios introduce significant uncertainty in determining contact points. Peters et al. \cite{SM} demonstrate that having a vehicle suspension system can complicate the localization of these contact points. These authors \cite{SM,norouzi2013statistical} also document that accurately determining contact points for a vehicle with a suspension system requires the use of suspension displacement sensors and/or a kinematic model of the suspension system, and accurately measuring the vehicle's real-time wheel forces requires expensive wheel force sensors. Due to these limitations, physics-driven methods can be impractical to implement and deploy.



\subsection{Learning-based Approaches} Learning-based metrics provide a way to simplify stability estimation by eliminating the need to localize contact points on the terrain, install expensive hardware, or use a vehicle dynamics model. In the context of this paper, traversability metrics and stability metrics will be considered interchangeable, as a traversability metric can be considered a quantitative or qualitative estimate of how statically or dynamically stable a UGV will be while traveling over different terrain types \cite{martin2018proprioceptive}. There are many extensive surveys covering traversabilty estimation methods \cite{papadakis2013terrain, borges2022survey, benrabah2024review, shu2024overview}; we discuss some commonly used learning-based metrics in this paper.

\subsubsection{Traversability Estimation}
Learning-based traversability estimation commonly leverages data, particularly from IMUs and wheel encoders, to analyze the vibration patterns experienced by UGVs as they traverse over various terrain types. These vibration signals have been examined in both the time and frequency domains. Early approaches employed traditional machine learning models for classification, such as Support Vector Machines (SVMs) \cite{bermudez2012performance}, Extreme Learning Machines (ELMs) \cite{vicente2015surface}, and basic multilayer perceptrons (MLPs) \cite{bai2019deep}. More recent research has enhanced classification performance through the application of deep learning architectures, including recurrent neural networks (RNNs) \cite{otte2016recurrent} and convolutional neural networks (CNNs) \cite{vulpi2021recurrent}. Additional methods incorporate force or tactile sensor data for terrain estimation using learning-based techniques \cite{wu2019tactile}, while others utilize acoustic signals generated by the UGV's interaction with the terrain \cite{fragoso2025acoustic}. A growing trend involves fusing data from multiple sources to construct multi-modal input representations, which are then processed using deep neural networks to improve terrain classification accuracy \cite{guo2024supervised, zhang2025terrain}. 

\subsubsection{Self-Supervised Near-to-far learning} 
Proprioceptive traversability estimation is frequently employed as a source of pseudo-ground truth labels for training exteroceptive-based traversability models in self-supervised learning frameworks for autonomous navigation \cite{bajracharya2009autonomous, zurn2020self, oliveira2021three, mayuku2021self, kahn2021badgr, chen2024terrain}. For instance, Bajracharya et al. \cite{bajracharya2009autonomous} used a Naive Bayes classifier trained on data,  including wheel encoders, bumper switches, IMU, GPS, and motor current sensors, to generate binary terrain labels (traversable vs. non-traversable). These labels were subsequently used to train a series of one-class SVMs to learn terrain classifications from stereo vision data. Zurn et al. \cite{zurn2020self} leveraged ground truth labels generated by a Siamese Encoder with Reconstruction Loss (SE-R), which used acoustic frequency-domain features to infer terrain types, for training a self-supervised semantic segmentation model based on camera image inputs. Kahn et al. \cite{kahn2021badgr} incorporated IMU-derived information about terrain bumpiness and collision likelihood into an LSTM-based model that predicts terrain traversability using exteroceptive inputs from cameras and LiDAR. This model was integrated into a reinforcement learning framework for smooth trajectory planning. Chen et al. \cite{chen2024terrain} use a one-dimensional CNN to process raw signals from IMU, wheel encoder, and motor current sensors to assess terrain traversability classes and uses these labels to improve results of a vision-based traversability estimation network that uses RGB and NIR image data. 

In contrast to approaches which adopt the near-to-far learning paradigm by using on-board feedback to supervise passive perception, our work proposes an inverse paradigm. Specifically, we employ vision-based sensors to estimate platform stability, using ArUco marker-based body displacement as a physically grounded and interpretable ground truth measure of stability. This vision-derived stability score serves as a reliable supervisory signal for training a neural network to predict platform stability directly from IMU data and measured velocity. The resulting stability estimation offers three key benefits: (1) it enables the quantitative assessment of terrain traversability along paths taken by an UGV, useful for evaluating traversability planning algorithms; (2) it provides high-quality ground truth labels for training self-supervised, exteroceptive-based traversability estimation models in a near-to-far learning framework; and (3) it supplies real-time stability feedback that can be integrated into stability-aware navigation systems, offering a robust fallback when exteroceptive estimators fail to accurately predict terrain characteristics. 

%% file: 3_Solution_Approach.tex
\section{Solution Approach}

In this section, we formulate the problem, introduce a stability metric (C3), and describe our overall learning pipeline for stability estimation.

\subsection{Problem Statement}
The objective of this work is to estimate a scalar \textit{stability score} for a wheeled mobile robot using onboard sensor data, specifically IMU and velocity measurements, captured over windowed time sequences. The stability score quantifies the robot's motion stability based on the pose deviation of the robot as the ground truth. This score captures how terrain-induced disturbances affect the robot's ability to operate safely, including the risk of rollover, which can compromise perception, actuation, and task performance. 

\subsection{Ground Truth Stability Measure}
\label{sec:c3}
We develop a camera-based stability metric, C3 score, for robot pose deviation to serve as a ground truth for a stability estimation method independent of exteroceptive sensing. To achieve this we use a camera rigidly attached to the UGV to track the position change of a static reference point (ArUco marker) in the environment. The ground truth measure captures movements of the center of the ArUco marker in the camera view across consecutive image frames as the UGV traverses the environment. The C3 score is adept at capturing the magnitude and direction of robot pose deviation on a single scale, efficiently combining into a single scalar value representing platform (in)stability. This metric provides an easy-to-compute supervisory signal for learning based stability prediction methods.

\begin{figure}[!htbp]
  \begin{center}
  \includegraphics[width=\linewidth]{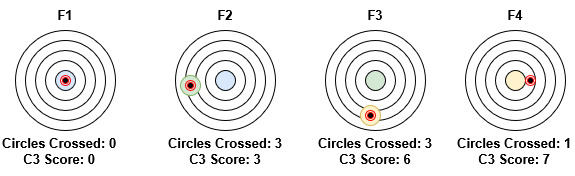}
  \caption{Count-Circle-Crossed (C3) Score Calculation based on ArUco Marker's Center Movement across Concentric Circles, using the Center from Previous ArUco Marker
  }
  \label{fig:ccc}
  \end{center}
\end{figure}

Our proposed metric, count-circles-crossed (C3), is designed to capture robot-terrain interactions over a sequence of frames. To compute the C3 score for a sequence of frames, we first compute the CC (circles-crossed) score for a single image. The CC score is computed in reference to the preceding frame. To compute the score we draw twenty concentric circles with linearly-increasing radii (2 to 40 pixel units radius) around the ArUco marker center in the previous image frame and calculate the number of circles jumped by the ArUco center in the current frame. The sum of CC values for all frames in a windowed sequence is called the C3 score for the sequence. Algorithm \ref{c3_algo} gives the psuedocode for C3 score calculation, where $F$ is an array of size $N$ of consecutive image frames. A visualization for this method is shown in Fig \ref{fig:ccc}.

\begin{algorithm}
  \caption{C3 Score Calculation}\label{c3_algo}
  \begin{algorithmic}[1]
    \Procedure{C3Score}{$F$}
      \State $C3S\gets 0$
      \For{\texttt{i = 2, ..., N}}
        \State $prevCenter\gets findArUcoCenter(F_{i-1})$
        \State $circles\gets generateCircles(prevCenter)$ 
        
        \Comment{Generate 20 concentric circles (linearly-increasing radii) around ArUco center in previous frame}
        
        \State $curCenter\gets findArUcoCenter(F_{i})$
       
        \Comment{Count number of circles between ArUco centers in previous and current frames}
        
        \State $CC\gets NumCirclesCrossed(curCenter, circles)$
        
         \Comment{$d_{ArUco}$: robot's current distance from ArUco}
         
         \Comment{$d_{Max}$: robot's longest travel distance in field}
         
        \State $ CC \gets CC \times (d_{ArUco}/d_{Max})$
        \State $C3S\gets  C3S+CC$
      \EndFor
      \State \Return $C3S$
    \EndProcedure
  \end{algorithmic}
\end{algorithm}

\begin{figure*}[!htbp]
\centering
\includegraphics[width=0.80\linewidth]{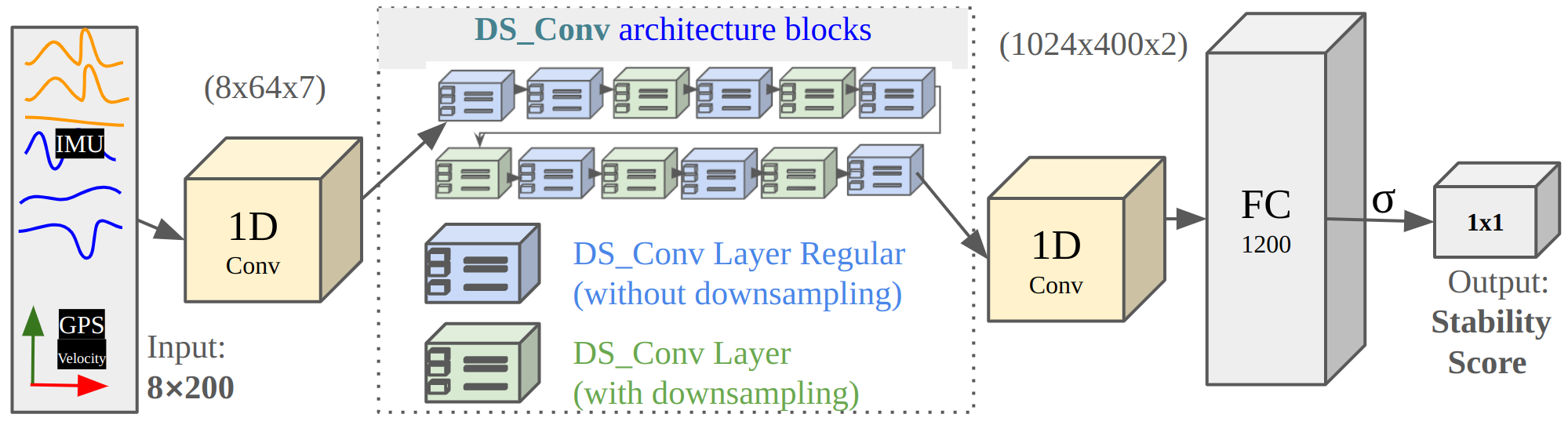}
\caption{Proposed Network Architecture for Learning the Stability Score}
\label{fig:san_net}
\end{figure*}

As the distance between the robot and the ArUco marker increses, the sensitivity of CC score to robot pose deviation increases. This is because the CC score is computed in pixel units. To correct this bias, we multiply the $CC$ score by a normalization factor (Algorithm \ref{c3_algo} step 8): \(CC = CC \times (d_{ArUco}/d_{Max}) \), where \( d_{{ArUco}} \) is the robot’s current distance from the ArUco marker and \( d_{{Max}} \) is the maximum distance observed during the trial. For ease of neural network training, we normalize the C3 score to lie in the range [0,1]. This is done by scaling the score with the maximum $C3$ score across all terrains ($C3_{max}$), as follows:
\(GT = \frac{C3}{C3_{max}} ~\epsilon ~[0,1]\).

\subsection{Learning Pipeline Overview}

We use the pretrained IMUNet model \cite{zeinali2024imunet} and fine-tune it to customize for stability prediction (Figure \ref{fig:san_net}). The pretrained network is trained on inertial motion data optimized for edge-device deployment and processes raw IMU sequences to estimate velocity or position. To adapt the model for our task, we remove the noise cancellation block. The architecture is then modified to accept an $8\times200$ input, corresponding to 6-axis IMU and velocity along robot's x and y axes. Our training set comprises of $~$1000 dataframes collected in 86 runs of a wheeled mobile robot across four different terrains (pavement, grass, dirt, and dirt with rocks) and three distinct velocity profiles ($\mathrm{0.5~ms^{-1}}$, $ \mathrm{1~ms^{-1}}$, and $\mathrm{1.5~ms^{-1}}$).

\subsubsection*{Training Strategy}
We train the network for $100$ epochs using a batch size of $32$ and the Adam optimizer with a learning rate of $1e^{-4}$. To prevent overfitting, we apply a dropout rate of $50\%$. Since the task involves predicting a continuous stability score, we use Mean Squared Error (MSE) as the loss function. We conduct all our experiments on a system with an NVIDIA RTX 3500 Ada GPU (12 GB VRAM) using CUDA 12.4. We collect sensor data from multiple terrains at varying speeds for training and reserve a separate, unseen terrain for evaluation. We use 15\% of the training dataset for validation. To evaluate our model, we use MSE as the primary evaluation metric.

%% file: 4_Data_Collection_and_Processing.tex
\section{Data Collection and Processing}
This section outlines our system and experimental design, ground truth stability score generation, and data preprocessing methods for model training.

\begin{center}
  \includegraphics[width=0.6\linewidth]{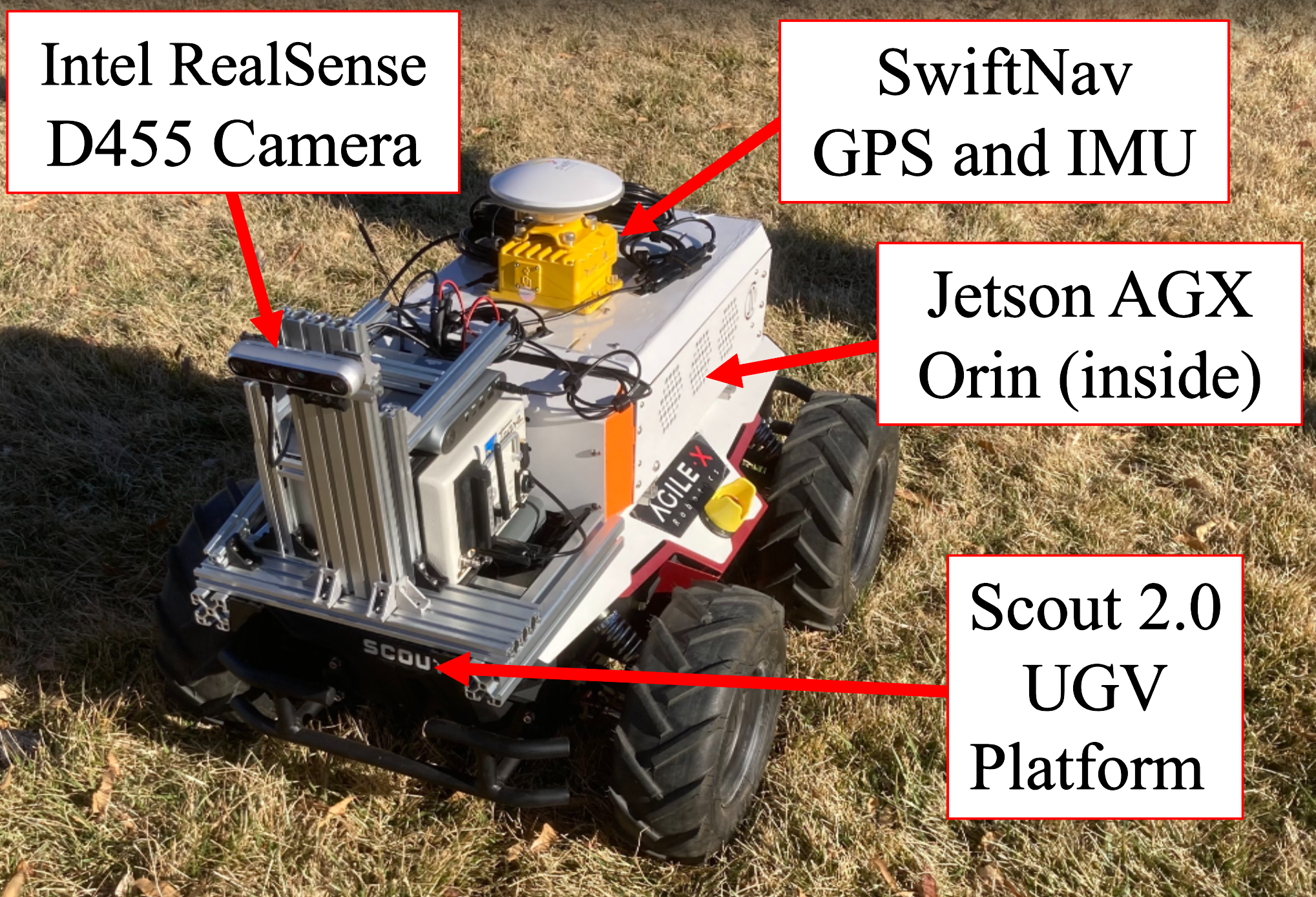}
  \captionof{figure}{System Design Setup: AgileX Scout 2.0 UGV Platform }
  \label{fig:scout}
\end{center}

\subsection{System Design}
We use an AgileX Scout 2.0 UGV for experimental data. The robot is equipped with a software stack designed in ROS2 providing localization and GPS based waypoint navigation capabilities. The onboard software is executed on an NVIDIA Jetson AGX Orin module. The robot is equipped with a SwiftNav Duro Inertial unit that provides RTK-GPS position data at 10Hz and IMU data at 200Hz. An Intel Realsense D455 camera provides images at 60Hz for ArUco marker detection and is rigidly mounted on the robot  (see Fig \ref{fig:scout}). 

\begin{figure*}[t]
    \centering
    \subfigure[]{\includegraphics[width=0.23\textwidth]{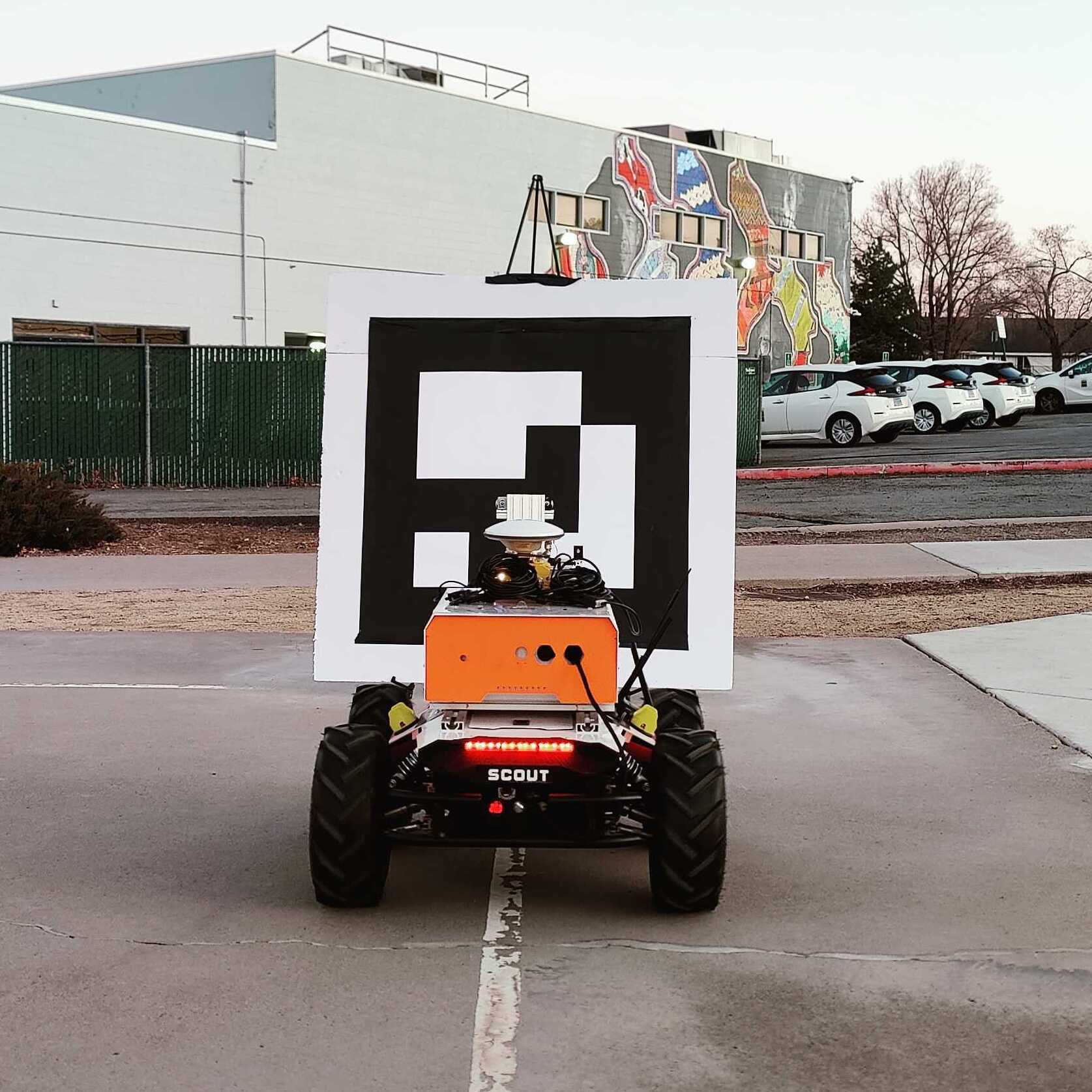}} 
    \subfigure[]{\includegraphics[width=0.23\textwidth]{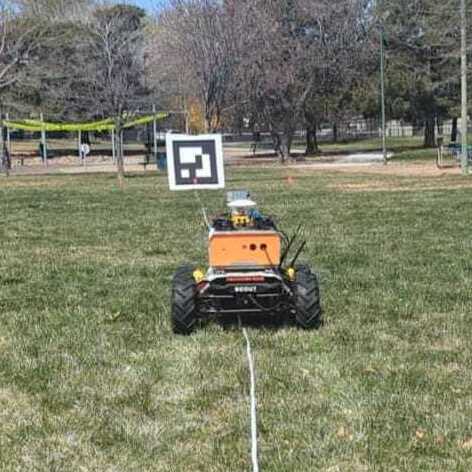}} 
    \subfigure[]{\includegraphics[width=0.23\textwidth]{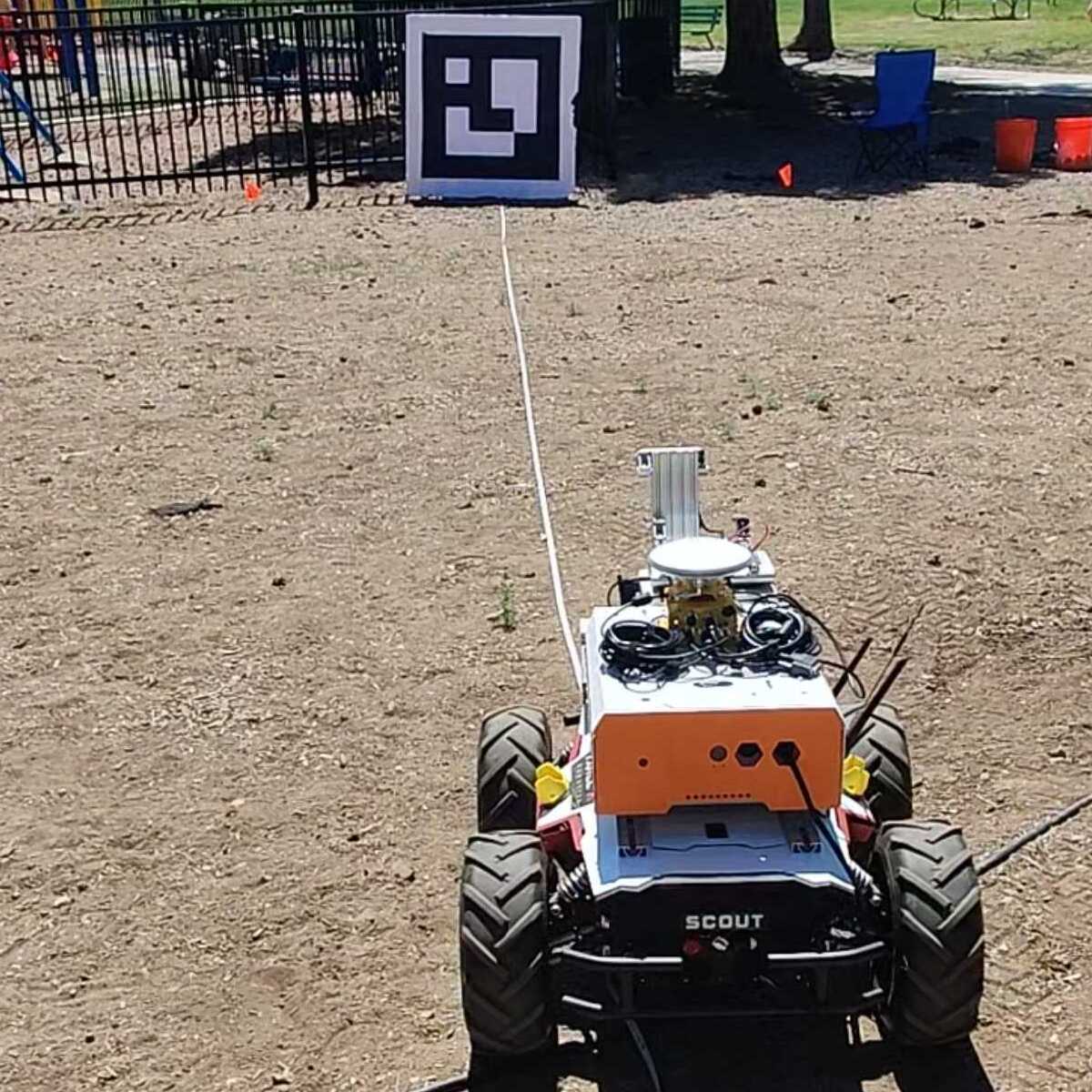}}
    \subfigure[]{\includegraphics[width=0.23\textwidth,trim=0 3mm 0 0,clip]{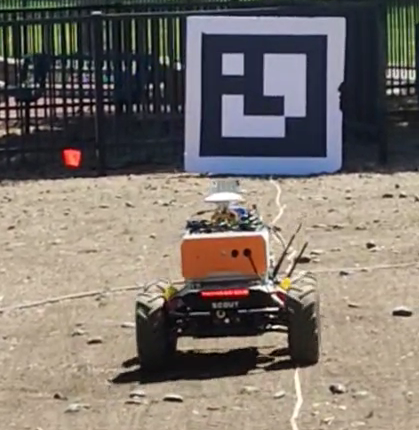}}
    \caption{Experiment setup with the Scout 2 
    facing an ArUco marker on (a) Pavement, (b) Grass, (c) Dirt, \& (d) Dirt with Rocks}
    \label{fig:experiments}
\end{figure*}

\begin{table*}[h]
\centering
\resizebox{\textwidth}{!}{
    \begin{tabular}{lcccccccc}
        \hline
        \textbf{Terrain\_Velocity} & \textbf{\# dataframes} & \textbf{Mean} & \textbf{Std dev} & \textbf{Min} & \textbf{25\%ile} & \textbf{median} & \textbf{75\%ile} & \textbf{Max} \\
        \hline 
        \hline 
        
        \multicolumn{8}{c}{\textbf{Training Data}} \\
        \hline 
        pavement\_0.5 & 123 & 0.000253 & 0.001719 & 0.000000 & 0.000000 & 0.000000 & 0.000000 & 0.016162 \\
        pavement\_1.0 & 82  & 0.000548 & 0.004274 & 0.000000 & 0.000000 & 0.000000 & 0.000000 & 0.038427 \\
        pavement\_1.5 & 115 & 0.006886 & 0.025879 & 0.000000 & 0.000000 & 0.000000 & 0.000000 & 0.184089 \\
        dirt\_0.5 & 241 & 0.015030 & 0.051415 & 0.000000 & 0.000000 & 0.000000 & 0.000000 & 0.564229 \\
        dirt\_1.0 & 116 & 0.109996 & 0.148408 & 0.000000 & 0.000000 & 0.074855 & 0.147032 & 0.673900 \\
        dirt\_1.5 & 106 & 0.238046 & 0.245232 & 0.000000 & 0.067926 & 0.157616 & 0.336529 & 1.000000 \\
        dirt+rocks\_0.5 & 357 & 0.032166 & 0.093634 & 0.000000 & 0.000000 & 0.000000 & 0.000000 & 0.700692 \\
        dirt+rocks\_1.0 & 166 & 0.215782 & 0.238787 & 0.000000 & 0.005403 & 0.136750 & 0.353867 & 0.999209 \\
        dirt+rocks\_1.5 & 47  & 0.338978 & 0.288741 & 0.000000 & 0.099221 & 0.265559 & 0.545717 & 0.972803 \\
        \hline
        \multicolumn{8}{c}{\textbf{Test Data (unseen terrain)}} \\
        \hline 
        grass\_0.5 & 181 & 0.000904 & 0.004865 & 0.000000 & 0.000000 & 0.000000 & 0.000000 & 0.055362 \\
        grass\_1.0 & 91  & 0.053647 & 0.056334 & 0.000000 & 0.002324 & 0.043401 & 0.076191 & 0.305136 \\
        grass\_1.5 & 175 & 0.130626 & 0.145167 & 0.000000 & 0.027450 & 0.081663 & 0.192160 & 0.804693 \\
        \hline
    \end{tabular}
    }
    \caption{Stability Score (Ground Truth) Statistics - Terrain Type with Velocity ($ms^{-1}$)}
    \label{tab:terrain_gt}
\end{table*}

\subsection{Experiment Design}
Experiments were conducted in four different terrain types: ``pavement", ``grass", ``dirt" and ``dirt with rocks" (See  Fig \ref{fig:experiments}). We conducted controlled trials for all of these terrains at different speeds (0.5, 1, and 1.5) $\mathrm{ms^{-1}}$, to capture data at varying levels of stability, as we expected higher speeds to lead to instability as the robot navigates through uneven terrain. Each trial began with the robot aligned to face an ArUco marker of size $\mathrm{1~m^2}$ for easy detection and visibility. During each trial, the robot moved forward towards the marker with a constant, positive linear velocity along the x-axis as input. ROSbags were used to collect IMU, GPS, and camera data for each trial.

\subsection{Preprocessing}
\label{sec:dataframe}
Sensor frame rates vary, so we preprocess the raw sensor data to create fixed-length inputs for the network. Specifically, we use 200 sequential IMU measurements as the reference and define a time window spanning from the first to the last IMU measurement. We use GPS co-ordinates to compute the velocity. Camera and GPS measurements within the window are tracked, with GPS data interpolated to match the IMU timestamps. This process yields a $8 \times 200$ \textit{dataframe}, where each dataframe contains six IMU features (angular velocities and accelerations in X, Y, Z) and two velocity features (in X and Y). We remove outliers from the collected data caused by unintended robot movements, such as during boot-up, stationary periods, or when the robot is aligning with the ArUco marker vector.

\begin{figure}[!hb]
    \centering
    \includegraphics[width=0.9\linewidth]{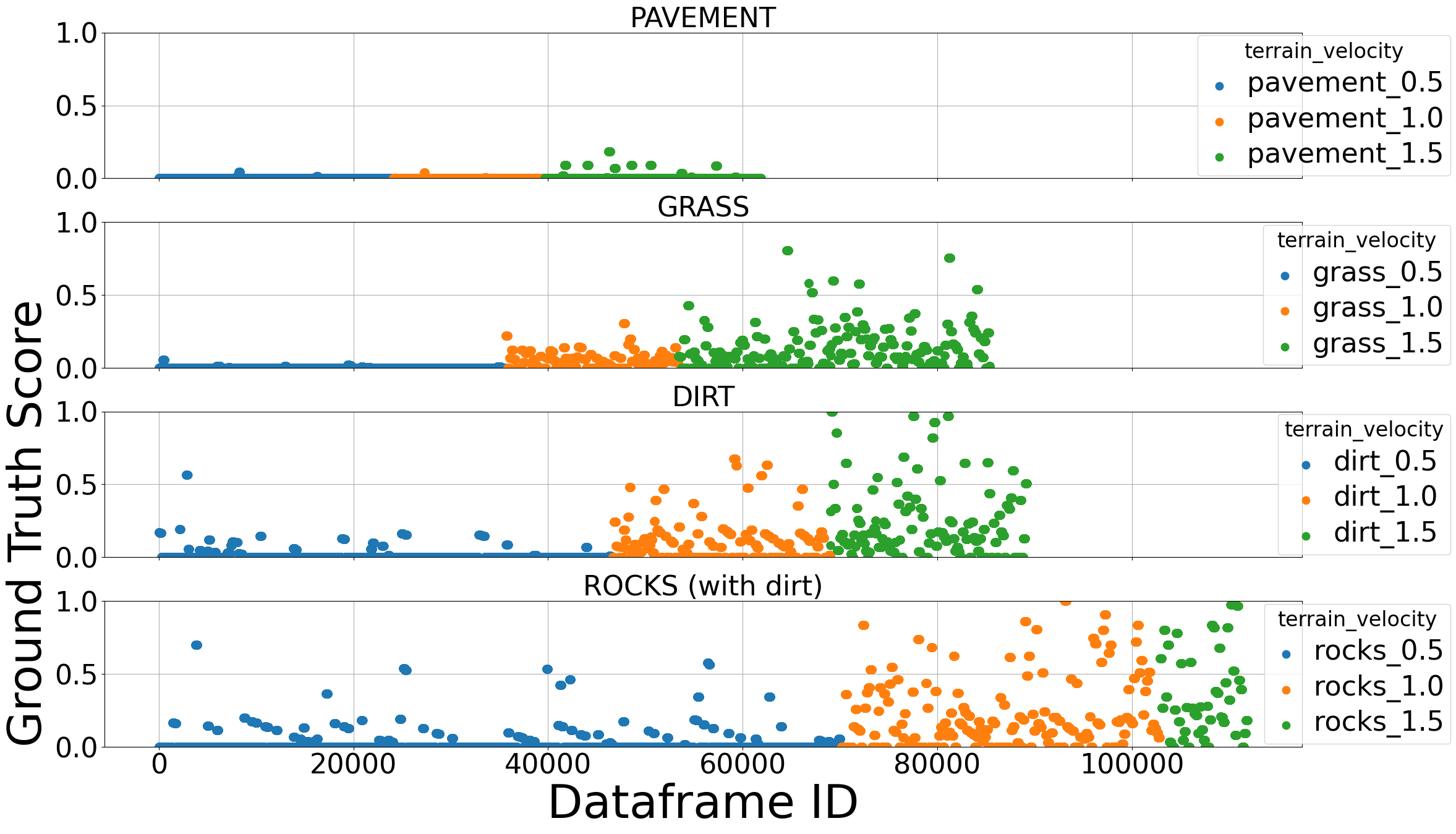}
    \caption{Normalized Ground Truth values for each terrain class. Data is sorted by input velocity (m/s) provided to the UGV.}
    \label{fig:terrain_gt}
\end{figure}
\subsection{Ground Truth Distribution and Dataset Balancing}
In Table~\ref{tab:terrain_gt}, we present the ground truth statistics of our collected dataset, which includes data from ``pavement", ``grass", ``dirt" and ``dirt with rocks" terrain types. The mean stability score exhibits a consistent increase as the terrain gets rougher ($\rm{rocks} >> \rm{dirt} >> \rm{grass} >> \rm{pavement}$) and input velocities increase. These ground truth scores support our hypothesis that terrain classes possess inherent instability and that the UGV’s velocity influences this instability, as evidenced by the magnitude of dynamic fluctuations observed. The score distributions further indicate that our ground truth scoring method effectively captures the instability characteristics across each terrain-velocity class.
However, the collected dataset shows an imbalance in terrain-velocity classes within the training data (see Figure \ref{fig:terrain_gt}). To address this, we sample an equal number of dataframes from each velocity level within each terrain type. 
Additionally, we apply stratified sampling by binning ground truth score values to ensure that both training and validation sets maintain a consistent distribution of ground truth scores.
This balanced distribution helps the model reduce bias, avoid overfitting to skewed subsets, and learn effectively across the full range of stability present in the training data. 

%% file: 5_Results.tex
\section{Results and Discussion}

\begin{figure}[h]
    \centering
    \includegraphics[width=0.6\linewidth]{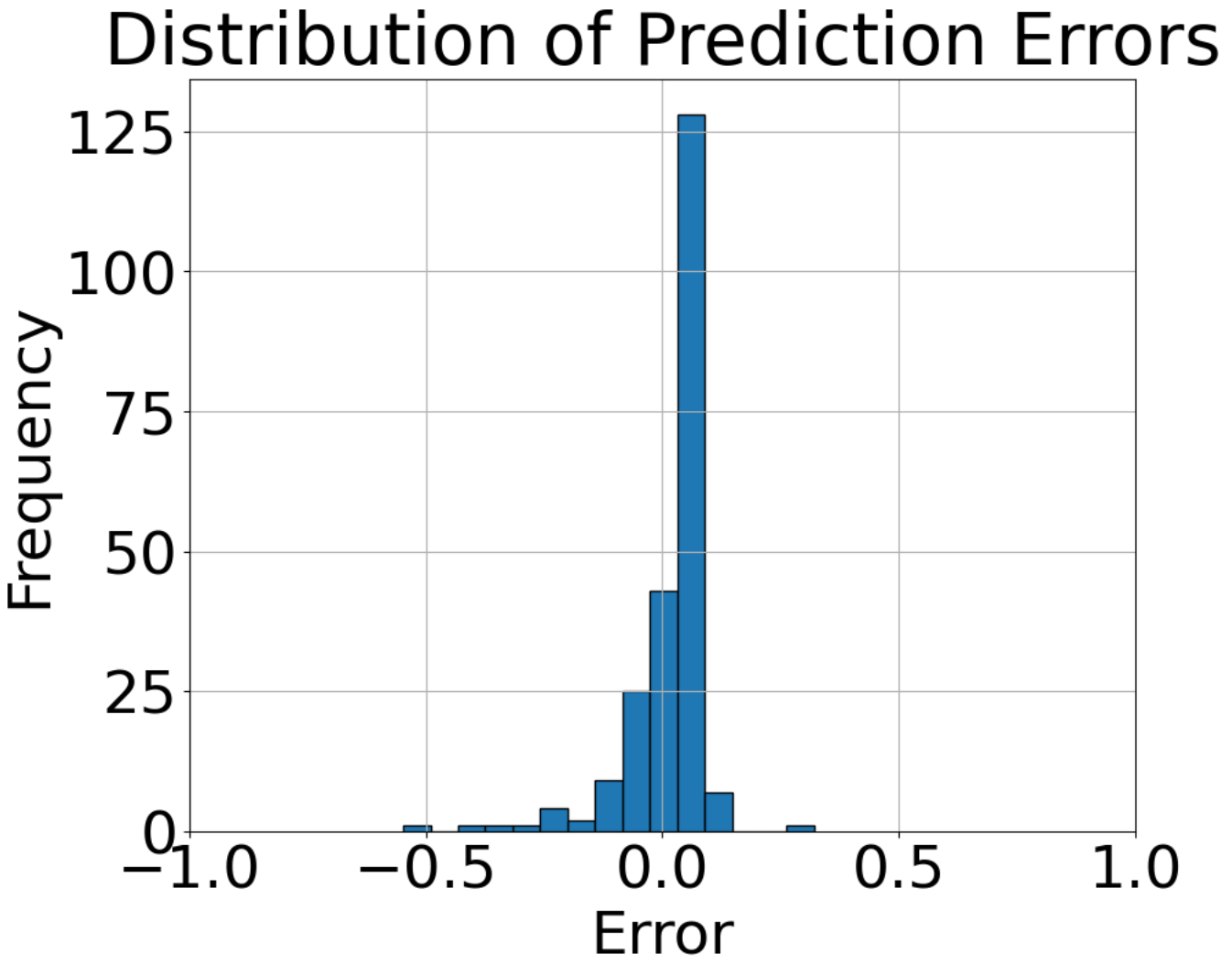}
    \caption{Distribution of Stability Score Prediction Errors}
    \label{fig:pred_error_dist}
\end{figure}

\begin{figure*}[]
\centering
  \centering
  \includegraphics[width=00.9\linewidth]{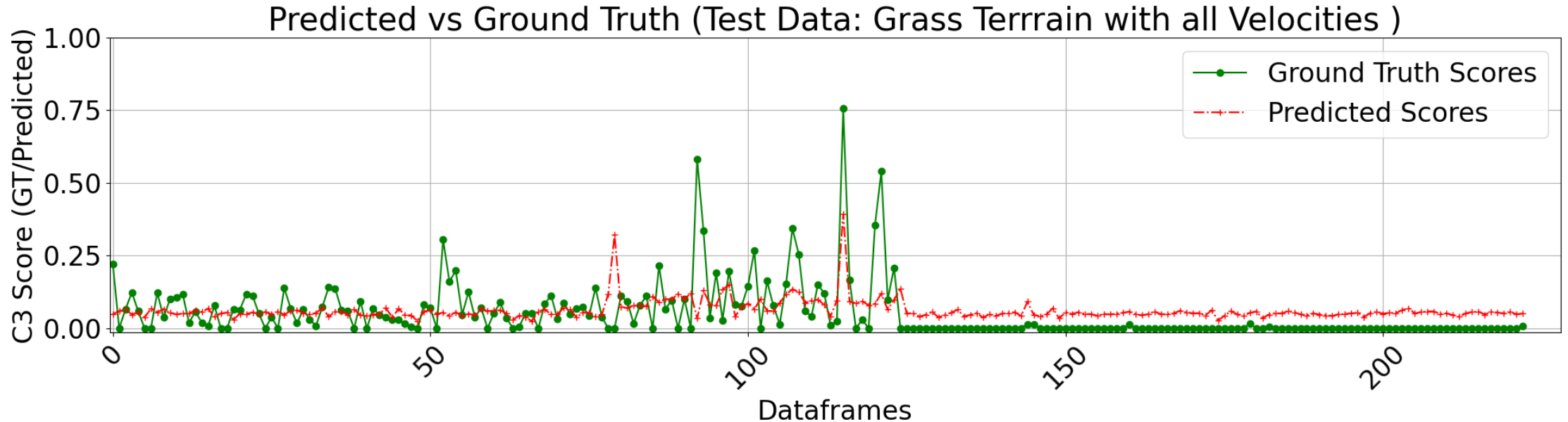}
  \caption{ Stability Score Prediction on Unseen Test Data 
  } 
  \label{fig:pred_result}
\end{figure*}

We evaluated the performance of our system on test data comprising terrain class ``grass", previously unseen by the model, across all velocity profiles. The model achieves an MSE of $0.0074$, indicating relatively low prediction errors within the C3 score range of zero to one. Fig.~\ref{fig:pred_error_dist} shows the distribution of prediction errors. Most errors are close to zero, showing the overall performance of our model. However, the distribution shows a small bias, indicating a tendency to overestimate instability. This is explained by higher percentage of high C3 score dataframes in the training data representing ``dirt" and ``dirt with rocks" terrain classes, as seen in Figure ~\ref{fig:pred_result}. Figure ~\ref{fig:pred_result} shows both the predicted and ground truth stability scores. The comparison reveals that the model struggles with accuracy particularly when the ground truth indicates higher instability.

Although the model is trained on non-grass terrain classes, it achieves reasonably low prediction errors and demonstrates the ability to predict stability across terrain types and velocity profiles. This suggests that IMU and velocity features capture terrain-agnostic information that shows limited generalization with current dataset and model architecture.

%% file: 6_Conclusion_And_FutureWork.tex
\section{Conclusion and Future Work}

In this work, we proposed a new camera-based stability measure that quantifies robot pose deviation using ArUco marker displacement. Initial experiments suggest that the C3 score provides a reliable and interpretable measure of wheeled robot stability. We observe that the C3 score increases as the robot traverses rougher terrains and operates at higher speeds, indicating that the metric captures terrain-robot interactions. The IMU-velocity based prediction method offers a promising direction for real-time stability estimation using onboard proprioceptive sensing and robot state information. It has wide-ranging downstream applications, including gating actuation and data collection, retrospective terrain labeling, and as a training signal for self-supervised traversability estimation and stability prediction.

As part of future work, we plan to expand the dataset across existing terrain-velocity classes to improve representation amongst classes and increase the percentage of dataframes with higher C3 scores. While early results offer evidence of the feasibility of IMU and velocity-based prediction, more rigorous evaluations and benchmarking are needed to fully understand the capabilities of the proposed stability estimation method. We plan to run ablation studies to better understand the specific contributions of different IMU channels and velocity components to model performance.

We will also evaluate and explore other neural architectures (time-series models) and training strategies. We will broaden the variance of our dataset by targeting new terrain-velocity combinations that lead to greater variability in the C3 score, especially those associated with unstable and dynamic robot motion. We will also introduce slope conditions (no slope, moderate, and steep) to increase exposure to tip-over scenarios. To validate the effectiveness and generalizability of our approach, we will compare it against classical terrain classification baselines that use statistical and frequency-domain features of IMU signals, such as \cite{vicente2015surface}. In addition, we will benchmark our method against well-established physics-based stability metrics, including Static Stability Margin (SSM) \cite{SSM}, Force-Angle Stability Margin (FASM) \cite{FASM}, and Stability Moment (SM) \cite{SM}.

%% file: acknowledgement.tex
\section*{Acknowledgement}
This work is supported by the NASA EPSCoR Program and the NV NASA Space Consortium as part of the project titled “ARC: Prospecting and Pre-Colonization of the Moon and Mars using Autonomous Robots with Human-in-the-Loop.”

%% file: _main_RSS_ROAR.bbl
\begin{thebibliography}{10}

\bibitem{SSM}
Robert~B McGhee and Andrew~A Frank.
\newblock \href{https://doi.org/10.1016/0025-5564(68)90090-4}{On the stability properties of quadruped creeping gaits}.
\newblock {\em Mathematical Biosciences}, 3:331--351, 1968.

\bibitem{ZMP}
Miomir Vukobratovi{\'c} and Branislav Borovac.
\newblock \href{https://doi.org/10.1142/S0219843604000083}{Zero-moment point—thirty five years of its life}.
\newblock {\em International journal of humanoid robotics}, 1(01):157--173, 2004.

\bibitem{ESM}
Dominic Messuri and Charles Klein.
\newblock \href{https://ieeexplore.ieee.org/abstract/document/1087012?casa_token=Q2w2VOM_uOUAAAAA:wN-n77AGvzeJxu_NfpTB8aFhzB5qvbWALd2YFFs7CA7mr3fTxo8yTO_J5CYul4bHec2PKsWN}{Automatic body regulation for maintaining stability of a legged vehicle during rough-terrain locomotion}.
\newblock {\em IEEE Journal on Robotics and Automation}, 1(3):132--141, 1985.

\bibitem{mayuku2021self}
Orighomisan Mayuku, Brian~W Surgenor, and Joshua~A Marshall.
\newblock \href{https://ieeexplore.ieee.org/abstract/document/9562029}{A self-supervised near-to-far approach for terrain-adaptive off-road autonomous driving}.
\newblock In {\em 2021 IEEE International Conference on Robotics and Automation (ICRA)}, pages 14054--14060. IEEE, 2021.

\bibitem{ryu2024evaluation}
Sijun Ryu, Jeeho Won, Hobyeung Chae, Hwa~Soo Kim, and TaeWon Seo.
\newblock \href{https://doi.org/10.1007/s12541-023-00912-6}{Evaluation criterion of wheeled mobile robotic platforms on grounds: A survey}.
\newblock {\em International Journal of Precision Engineering and Manufacturing}, 25(3):675--686, 2024.

\bibitem{ryu2024simulation}
Sijun Ryu, Jeeho Won, and TaeWon Seo.
\newblock \href{https://doi.org/10.1016/j.robot.2024.104749}{Simulation study on four-wheeled mobile robot mechanisms using various performance criteria}.
\newblock {\em Robotics and Autonomous Systems}, 179:104749, 2024.

\bibitem{FASM}
EG~Papadopoulos and Daniel~A Rey.
\newblock \href{https://doi.org/10.1109/ROBOT.1996.509185}{A new measure of tipover stability margin for mobile manipulators}.
\newblock In {\em Proceedings of IEEE International Conference on Robotics and Automation}, volume~4, pages 3111--3116. IEEE, 1996.

\bibitem{MHS}
Shady Ali, A~Moosavian, and Khalil Alipour.
\newblock \href{https://ieeexplore.ieee.org/abstract/document/4018846}{Stability evaluation of mobile robotic systems using moment-height measure}.
\newblock In {\em 2006 IEEE conference on robotics, automation and mechatronics}, pages 1--6. IEEE, 2006.

\bibitem{SM}
Steven~C Peters and Karl Iagnemma.
\newblock \href{https://doi.org/10.1080/00423110802344636}{Stability measurement of high-speed vehicles}.
\newblock {\em Vehicle System Dynamics}, 47(6):701--720, 2009.

\bibitem{norouzi2013statistical}
Mohammad Norouzi, Jaime~Valls Miro, and Gamini Dissanayake.
\newblock \href{https://ieeexplore.ieee.org/abstract/document/6630575}{A statistical approach for uncertain stability analysis of mobile robots}.
\newblock In {\em 2013 IEEE International Conference on Robotics and Automation}, pages 191--196. IEEE, 2013.

\bibitem{martin2018proprioceptive}
Steven~C Martin.
\newblock {\em \href{https://eprints.qut.edu.au/117074/}{Proprioceptive sensing of traversability for long-term navigation of robots}}.
\newblock PhD thesis, Queensland University of Technology, 2018.

\bibitem{papadakis2013terrain}
Panagiotis Papadakis.
\newblock \href{https://doi.org/10.1016/j.engappai.2013.01.006}{Terrain traversability analysis methods for unmanned ground vehicles: A survey}.
\newblock {\em Engineering Applications of Artificial Intelligence}, 26(4):1373--1385, 2013.

\bibitem{borges2022survey}
Paulo~VK Borges, Thierry Peynot, Sisi Liang, Bilal Arain, Matthew Wildie, Melih~G Minareci, Serge Lichman, Garima Samvedi, Inkyu Sa, Nicolas Hudson, et~al.
\newblock \href{https://ieeexplore.ieee.org/abstract/document/10876033}{A survey on terrain traversability analysis for autonomous ground vehicles: Methods, sensors, and challenges}.
\newblock {\em Field Robotics}, 2:1567--1627, 2022.

\bibitem{benrabah2024review}
Mohamed Benrabah, Charifou Orou~Mousse, Elie Randriamiarintsoa, Roland Chapuis, and Romuald Aufr{\`e}re.
\newblock \href{https://doi.org/10.3390/s24061909}{A Review on Traversability Risk Assessments for Autonomous Ground Vehicles: Methods and Metrics}.
\newblock {\em Sensors}, 24(6):1909, 2024.

\bibitem{shu2024overview}
Yongjie Shu, Linwei Dong, Jianfeng Liu, Cheng Liu, and Wei Wei.
\newblock \href{https://doi.org/10.1002/rob.22461}{Overview of Terrain Traversability Evaluation for Autonomous Robots}.
\newblock {\em Journal of Field Robotics}, 2024.

\bibitem{bermudez2012performance}
Fernando L~Garcia Bermudez, Ryan~C Julian, Duncan~W Haldane, Pieter Abbeel, and Ronald~S Fearing.
\newblock \href{https://ieeexplore.ieee.org/abstract/document/6386243?casa_token=ZpwKLqfLF9oAAAAA:kX9QSKWqfMb18JZXaMI11-Dp3C7K4mSoAV75Paf88P8Fx-5PCYHMDaXGXvkQKf5HTybugMbh}{Performance analysis and terrain classification for a legged robot over rough terrain}.
\newblock In {\em 2012 IEEE/RSJ international conference on intelligent robots and systems}, pages 513--519. IEEE, 2012.

\bibitem{vicente2015surface}
Alexandre Vicente, Jindong Liu, and Guang-Zhong Yang.
\newblock \href{https://ieeexplore.ieee.org/abstract/document/7353480}{Surface classification based on vibration on omni-wheel mobile base}.
\newblock In {\em 2015 IEEE/RSJ International Conference on Intelligent Robots and Systems (IROS)}, pages 916--921. IEEE, 2015.

\bibitem{bai2019deep}
Chengchao Bai, Jifeng Guo, Linli Guo, and Junlin Song.
\newblock \href{https://doi.org/10.3390/s19143102}{Deep multi-layer perception based terrain classification for planetary exploration rovers}.
\newblock {\em Sensors}, 19(14):3102, 2019.

\bibitem{otte2016recurrent}
Sebastian Otte, Christian Weiss, Tobias Scherer, and Andreas Zell.
\newblock \href{https://ieeexplore.ieee.org/abstract/document/7487778?casa_token=u9t7rY5OY8EAAAAA:f_a34MyV-NRHyawCiHQEuh1xtrXQ_hXvWadjClWz8QKmhthR7Y_-VOOzIbX3pyv5Yqiprxb1}{Recurrent neural networks for fast and robust vibration-based ground classification on mobile robots}.
\newblock In {\em 2016 IEEE International Conference on Robotics and Automation (ICRA)}, pages 5603--5608. IEEE, 2016.

\bibitem{vulpi2021recurrent}
Fabio Vulpi, Annalisa Milella, Roberto Marani, and Giulio Reina.
\newblock \href{https://doi.org/10.1016/j.jterra.2020.12.002}{Recurrent and convolutional neural networks for deep terrain classification by autonomous robots}.
\newblock {\em Journal of Terramechanics}, 96:119--131, 2021.

\bibitem{wu2019tactile}
X~Alice Wu, Tae~Myung Huh, Aaron Sabin, Srinivasan~A Suresh, and Mark~R Cutkosky.
\newblock \href{https://ieeexplore.ieee.org/abstract/document/8823987}{Tactile sensing and terrain-based gait control for small legged robots}.
\newblock {\em IEEE Transactions on Robotics}, 36(1):15--27, 2019.

\bibitem{fragoso2025acoustic}
Anthony~T Fragoso.
\newblock \href{https://doi.org/10.1016/j.jterra.2024.101028}{Acoustic winter terrain classification for offroad autonomous vehicles}.
\newblock {\em Journal of Terramechanics}, 117:101028, 2025.

\bibitem{guo2024supervised}
Jianbo Guo, Shuai Wang, Yiwei Mao, Guoqiang Wang, Guohua Wu, Yewei Wu, and Zhengbin Liu.
\newblock \href{https://doi.org/10.1016/j.compag.2024.108791}{Supervised learning study on ground classification and state recognition of agricultural robots based on multi-source vibration data fusion}.
\newblock {\em Computers and Electronics in Agriculture}, 219:108791, 2024.

\bibitem{zhang2025terrain}
Yinglong Zhang, Baoru Huang, Meng Hong, Chao Huang, Guan Wang, and Min Guo.
\newblock \href{https://doi.org/10.3390/electronics14061231}{A Terrain Classification Method for Quadruped Robots with Proprioception}.
\newblock {\em Electronics}, 14(6):1231, 2025.

\bibitem{bajracharya2009autonomous}
Max Bajracharya, Andrew Howard, Larry~H Matthies, Benyang Tang, and Michael Turmon.
\newblock \href{https://doi.org/10.1002/rob.20269}{Autonomous off-road navigation with end-to-end learning for the LAGR program}.
\newblock {\em Journal of Field Robotics}, 26(1):3--25, 2009.

\bibitem{zurn2020self}
Jannik Z{\"u}rn, Wolfram Burgard, and Abhinav Valada.
\newblock \href{https://ieeexplore.ieee.org/abstract/document/9247267}{Self-supervised visual terrain classification from unsupervised acoustic feature learning}.
\newblock {\em IEEE Transactions on Robotics}, 37(2):466--481, 2020.

\bibitem{oliveira2021three}
Felipe~G Oliveira, Armando~A Neto, David Howard, Paulo Borges, Mario~FM Campos, and Douglas~G Macharet.
\newblock \href{https://doi.org/10.1007/s10846-020-01304-y}{Three-dimensional mapping with augmented navigation cost through deep learning}.
\newblock {\em Journal of Intelligent \& Robotic Systems}, 101(3):50, 2021.

\bibitem{kahn2021badgr}
Gregory Kahn, Pieter Abbeel, and Sergey Levine.
\newblock \href{https://ieeexplore.ieee.org/abstract/document/9345970}{Badgr: An autonomous self-supervised learning-based navigation system}.
\newblock {\em IEEE Robotics and Automation Letters}, 6(2):1312--1319, 2021.

\bibitem{chen2024terrain}
Hsiao-Yu Chen, I-Chen Sang, William~R Norris, Ahmet Soylemezoglu, and Dustin Nottage.
\newblock \href{https://doi.org/10.1016/j.compag.2024.109539}{Terrain classification method using an NIR or RGB camera with a CNN-based fusion of vision and a reduced-order proprioception model}.
\newblock {\em Computers and Electronics in Agriculture}, 227:109539, 2024.

\bibitem{zeinali2024imunet}
Behnam Zeinali, Hadi Zanddizari, and Morris~J Chang.
\newblock \href{https://ieeexplore.ieee.org/document/10480886}{IMUNet: Efficient Regression Architecture for Inertial IMU Navigation and Positioning}.
\newblock {\em IEEE Transactions on Instrumentation and Measurement}, 2024.

\end{thebibliography}
